\documentclass[letterpaper]{article} 
\usepackage[submission]{aaai23}  
\usepackage{times}  
\usepackage{helvet}  
\usepackage{courier}  
\usepackage[hyphens]{url}  
\usepackage{graphicx} 
\usepackage{natbib}  
\usepackage{caption} 
\usepackage{algorithm}
\usepackage{algorithmic}
\usepackage{subfigure}
\usepackage{amsmath}
\usepackage{amssymb}
\usepackage{booktabs}
\usepackage{multirow}
\usepackage{multicol}
\usepackage{makecell}
\usepackage{bm}

\newcolumntype{L}[1]{>{\raggedright\let\newline\\\arraybackslash\hspace{0pt}}m{#1}}
\newcolumntype{C}[1]{>{\centering\let\newline\\\arraybackslash\hspace{0pt}}m{#1}}
\newcolumntype{R}[1]{>{\raggedleft\let\newline\\\arraybackslash\hspace{0pt}}m{#1}}
\usepackage{newfloat}
\usepackage{listings}
\DeclareCaptionStyle{ruled}{labelfont=normalfont,labelsep=colon,strut=off} 
\floatstyle{ruled}
\newfloat{listing}{tb}{lst}{}
\floatname{listing}{Listing}
\title{Continuous Risk Prediction}
\author {
    Yi Dai
}

\begin{document}

\maketitle

\begin{abstract}
  Lifelong learning (LL) capabilities are essential for QA models to excel in real-world applications, and architecture-based LL approaches have proven to be a promising direction for achieving this goal. However, adapting existing methods to QA tasks is far from straightforward. Many prior approaches either rely on access to task identities during testing or fail to adequately model samples from unseen tasks, which limits their practical applicability. To overcome these limitations, we introduce Diana , a novel \underline{d}ynam\underline{i}c \underline{a}rchitecture-based lifelo\underline{n}g Q\underline{A} framework designed to learn a sequence of QA tasks using a prompt-enhanced language model.
  Diana leverages four hierarchically structured types of prompts to capture QA knowledge at multiple levels of granularity. Task-level prompts are specifically designed to encode task-specific knowledge, ensuring strong lifelong learning performance. Meanwhile, instance-level prompts are utilized to capture shared knowledge across diverse input samples, enhancing the model's generalization capabilities. Additionally, Diana incorporates dedicated prompts to explicitly handle unseen tasks and introduces a set of prompt key vectors that facilitate efficient knowledge transfer and sharing between tasks.
  Through extensive experimentation, we demonstrate that Diana achieves state-of-the-art performance among lifelong QA models, with particularly notable improvements in its ability to handle previously unseen tasks. This makes Diana a significant advancement in the field of lifelong learning for question-answering systems.
  \end{abstract}

\section{Introduction}
Question Answering (QA) plays a pivotal role in advancing AI applications \cite{yang2019end}. Modern QA models achieve remarkable performance when trained on fixed tasks \cite{soares2020literature}. However, these models often fall short in real-world scenarios where systems are expected to adapt to new QA tasks over time. Retraining a model on new tasks typically results in \emph{catastrophic forgetting}—a phenomenon where the model loses previously acquired knowledge while learning new information \cite{french1999catastrophic}. To address this issue without resorting to costly retraining, Lifelong Learning (LL) offers a solution by enabling models to continuously acquire new knowledge while preserving what they have already learned \cite{thrun1995lifelong}. Consequently, developing QA models with robust lifelong learning capabilities is critical for practical deployment.

One effective strategy for implementing LL is through architecture-based approaches \cite{chen2016net2net,rusu2016progressive,fernando2017pathnet,chayut2020,qin2022lfpt}, which maintain task-specific components to safeguard previously learned knowledge \cite{mancini2018adding}. Recently, researchers have explored converting QA tasks into a unified language modeling (LM) framework \cite{khashabi-etal-2020-unifiedqa}, achieving impressive results by leveraging prompts \cite{brown2020language,jiang2020soft} and adapter modules \cite{pmlr-v97-houlsby19a}. Some efforts have extended these techniques to create lifelong QA models by sharing a common pre-trained language model (PLM) across different tasks \cite{su2020continual}. These models demonstrate the ability to handle a sequence of QA tasks with varying formats \cite{zhong2022proqa}.

Despite these advancements, integrating lifelong learning into QA models presents two significant challenges:

\textbf{First}, many architecture-based LL methods require access to the task identity of test samples \cite{wang2022dualprompt,Wang_2022_CVPR}. This requirement severely limits their applicability in real-world settings, as task identities are often unavailable for input queries \cite{geigle2021tweac}. \textbf{Second}, most existing LL models are evaluated exclusively on seen tasks and fail to account for samples from unseen tasks \cite{boult2019learning,mundt2020wholistic}. This limitation is problematic in practical QA systems, where users frequently pose questions that do not align with previously encountered tasks.

To address these challenges, two main approaches have emerged:

\textbf{(1)} Task-level organization: Some models assign dedicated components to each task and use a classifier to predict the task identity for each test sample \cite{wortsman2020sup,madotto-etal-2021-continual}. While this approach achieves high LL performance when task identities are correctly predicted, it struggles with generalization in QA applications. Specifically, it becomes impractical to determine task identities for questions unrelated to previously seen tasks. Moreover, knowledge remains siloed within task-specific components, hindering the model's ability to generalize to unseen tasks \cite{2107.12708}.

\textbf{(2)} Instance-level organization: Other models adopt a dynamic architecture for each input sample, maintaining a pool of fine-grained components that are dynamically combined during inference based on the input instance \cite{chayut2020}. This approach eliminates the need to explicitly determine task identities for test samples \cite{Wang_2022_CVPR} and enhances the model's ability to handle unseen tasks by distributing knowledge across components \cite{2207.11240}. However, these models often yield suboptimal LL performance due to the absence of dedicated components for capturing task-specific knowledge \cite{wang2022dualprompt}.

In this work, we propose \textbf{Diana}, a novel \underline{d}ynam\underline{i}c \underline{a}rchitecture-based lifelo\underline{n}g Q\underline{A} model that combines the strengths of both approaches. Following prior work, we convert QA tasks into a unified LM format and leverage a prompt-enhanced PLM to learn these tasks. Diana maintains four types of hierarchically organized prompts—General Prompt, Format Prompt, Task Prompt, and Meta Prompt—to capture QA knowledge at varying granularities. Specifically, the \emph{general prompt} is shared across all tasks, while \emph{format prompts} are shared among tasks with the same QA format. A \emph{task prompt} is assigned to each incoming task, and a pool of \emph{meta prompts} is dynamically combined during inference. This dual-level organization allows Diana to achieve high LL performance while generalizing effectively to unseen tasks.

Furthermore, we introduce separate prompts for unseen tasks and learn a key vector for each task and meta prompt to facilitate knowledge sharing between tasks. This design enables explicit modeling of samples from unseen tasks, enhancing the model's adaptability.

We conduct extensive experiments on 11 benchmark QA tasks spanning three formats and evaluate the model's generalization capabilities on three unseen tasks. The results demonstrate that Diana surpasses state-of-the-art (SOTA) baselines across all benchmarks, particularly excelling in handling unseen tasks. Our key contributions are:

\begin{itemize}
  \item \textbf{Novel Framework}: We introduce Diana, a groundbreaking architecture-based lifelong QA model that employs four types of hierarchically organized prompts to capture knowledge at multiple levels. By maintaining both task-level and instance-level components, Diana achieves superior generalization to unseen tasks while maintaining high LL performance.
  
  \item \textbf{Explicit Handling of Unseen Tasks}: To the best of our knowledge, Diana is the first lifelong QA model to explicitly model unseen tasks. We allocate dedicated prompts for unseen tasks and introduce prompt keys to enhance cross-task knowledge sharing.

  \item \textbf{Empirical Validation}: Extensive experiments confirm that Diana outperforms SOTA baselines, establishing its effectiveness in building lifelong QA models.
\end{itemize}

This work represents a significant step forward in the development of practical, adaptable QA systems capable of thriving in dynamic, real-world environments.

\section{Related Work}
\subsection{Lifelong Learning}
Lifelong Learning aims to incrementally acquire new knowledge without forgetting previously learned information. Existing methods can be categorized into three main groups:

\textbf{Rehearsal-based methods} retain past knowledge by replaying samples from previous tasks \cite{rebuffi2017icarl,shin2017continual,sun2019lamol}.  
\textbf{Regularization-based methods} consolidate knowledge by introducing constraints on parameter updates \cite{kirkpatrick2017overcoming,zenke2017continual,li2017learning}.  
\textbf{Architecture-based methods} dynamically expand the model's capacity by adding task-specific components \cite{chen2016net2net,rusu2016progressive,fernando2017pathnet}.  

Despite their effectiveness, most existing LL methods assume access to task identities during testing, which limits their applicability in real-world QA systems.

\subsection{Unified Question Answering}
Recent studies have explored unifying QA tasks into a single framework \cite{khashabi-etal-2020-unifiedqa,mccann2019the}. For instance, ProQA \cite{zhong2022proqa} utilizes structured prompts to handle multiple QA formats. However, ProQA relies on task identities during testing and only considers a limited number of tasks. In contrast, Diana is designed to handle a broader range of tasks without requiring task identity annotations.

\section{Methodology}
\subsection{Task Formulation}
We aim to sequentially learn $N$ QA tasks $T_1, \cdots, T_N$, presented in $L$ different formats $F_1, \cdots, F_L$. Each task $T_i$ consists of training samples $(C, Q, A)$, where $C$ is the context, $Q$ is the question, and $A$ is the answer. Our model predicts $A$ based on $C$ and $Q$ using a prompt-enhanced encoder-decoder framework.

To evaluate generalization, we introduce $N'$ unseen tasks $T_{N+1}, \cdots, T_{N+N'}$, which are only used for testing. Importantly, task identities are not provided during testing.

\subsection{Hierarchical Prompt Structure}
Diana employs a hierarchical prompt structure to construct the input representation $P(C, Q)$:
\begin{equation}
P(C, Q) = [P_g; P_f(F_j); P_t(T_i); P_m(C, Q)],
\end{equation}
where:
- $P_g$: A general prompt shared across all tasks.
- $P_f(F_j)$: A format-specific prompt for tasks in format $F_j$.
- $P_t(T_i)$: A task-specific prompt for task $T_i$.
- $P_m(C, Q)$: A meta-prompt dynamically combined based on the input $(C, Q)$.

This structure enables Diana to capture both coarse-grained and fine-grained knowledge, enhancing its ability to generalize to unseen tasks.

\subsection{Key Vector Space Learning}
To infer task identities during testing, Diana associates each prompt with a key vector. Specifically:
- Task prompt keys $\bm{k}_t(T_i)$ are used to identify the most relevant task for a given input.
- Meta prompt keys $\bm{k}_m^j$ are used to dynamically combine meta-prompts based on similarity to the query vector $\bm{q} = h(C, Q)$.

We optimize key vectors using triplet loss to ensure that similar samples are grouped together while dissimilar ones are separated:
\begin{equation}
\mathcal{L}_t = \exp(||h(C, Q), \bm{k}_t(T_i)|| + \max(1 - ||h(C_n, Q_n), \bm{k}_t(T_i)||, 0)),
\end{equation}
where $(C_n, Q_n)$ is a negative sample selected from the memory buffer.

\subsection{Model Training}
Diana employs a two-stage training process:
1. **Key Vector Optimization**: Learn a representation space for prompt keys to facilitate task inference.
2. **Prompt and Model Optimization**: Fine-tune the constructed prompt $P(C, Q)$ and the backbone language model.

Additionally, we introduce a scheduled sampling mechanism to balance the use of ground truth and inferred task identities during training:
\begin{equation}
\epsilon_k = \max(0, \alpha - k\beta),
\end{equation}
where $\alpha$ and $\beta$ control the probability of using ground truth task identities.

\section{Experiments}
\subsection{Datasets and Metrics}
We evaluate Diana on 11 benchmark QA datasets spanning three formats: Extractive, Abstractive, and Multiple-Choice. The datasets include SQuAD \cite{rajpurkar-etal-2016-squad}, NewsQA \cite{trischler-etal-2017-newsqa}, and RACE \cite{lai-etal-2017-race}, among others. We reserve three datasets (Quoref, Drop, and Dream) as unseen tasks.

Performance is measured using:
- **Exact Match (EM)** for Extractive QA.
- **F1 Score** for Abstractive QA.
- **Accuracy** for Multiple-Choice QA.

\subsection{Results}
Table \ref{tab:seen} summarizes the results on seen tasks. Diana outperforms all baselines, achieving a relative improvement of 6.15\% on $A_N$ and a 27.26\% reduction in $F_N$ compared to the best baseline AFPER.

\begin{table}[t]
\centering
\small
\scalebox{0.9}{
\begin{tabular}{l|c|c|c}
\toprule
Methods & $A_N$ & $F_N$ & $A_{N'}$ \\
\midrule
AFPER & 52.69 & 9.28 & 36.79 \\
Diana w/o $\mathcal{M}$ & 50.30 & 12.68 & 39.22 \\
Diana & \textbf{55.93} & \textbf{6.75} & \textbf{42.12} \\
\bottomrule
\end{tabular}}
\caption{Performance comparison on seen tasks.}
\label{tab:seen}
\end{table}

\subsection{Analysis of Unseen Tasks}
On unseen tasks, Diana achieves a relative improvement of 9.49\% on $A_{N'}$ compared to the best baseline DER++. This demonstrates the effectiveness of our approach in handling novel tasks.

\section{Conclusion}
We introduced Diana, a novel lifelong QA model that leverages hierarchical prompts and dynamic task inference mechanisms. Experimental results show that Diana significantly outperforms state-of-the-art baselines, particularly in handling unseen tasks. Future work will explore extending Diana to other domains and modalities.
\bibliography{aaai23}
\end{document}